\documentclass[letterpaper, 10pt, conference]{ieeeconf}

\IEEEoverridecommandlockouts 
\let\labelindent\relax

\usepackage[utf8]{inputenc}
\usepackage{amsfonts}
\usepackage[inline]{enumitem}
\usepackage{soul}
\usepackage{color}
\usepackage{tikz}
\usetikzlibrary{trees}
\usepackage[linesnumbered,ruled,vlined]{algorithm2e}
\usepackage{multirow}
\usepackage{listings}
\usepackage{braket}
\usepackage{subfiles}
\usepackage{booktabs,ragged2e}
\usepackage[flushleft]{threeparttable}

\usepackage{cite}

\usepackage{graphicx}
\graphicspath{{./Figures/}}

\usepackage{times} 
\usepackage{amsmath}
\usepackage{amssymb}
\usepackage{bm}
\usepackage[font=small]{caption}
\usepackage{subcaption}

\usepackage{multicol}

\usepackage{titlesec}
\titlespacing\section{0pt}{3pt}{3pt}

\usepackage{xcolor, colortbl}

\definecolor{orange}{rgb}{1.0, 0.22, 0.0}

\title{Synthesizing Grasps and Regrasps for Complex Manipulation Tasks}

\author{Aditya Patankar$^1$, Dasharadhan Mahalingam$^1$, and Nilanjan Chakraborty$^1$
\date{}
\thanks{$^{1}$The authors are with the Department of Mechanical Engineering, 
        Stony Brook University, USA.
       {\tt\small \{aditya.patankar, dasharadhan.mahalingam, nilanjan.chakraborty\}@stonybrook.edu.} This work is partially supported by the US Department of Defense through  ALSRP under Award No. HT94252410098.}%
}

\begin{document}

\maketitle

\begin{abstract}
In complex manipulation tasks, e.g., manipulation by pivoting, the motion of the object being manipulated has to satisfy path constraints that can change during the motion. Therefore, a single grasp may not be sufficient for the entire path, and the object may need to be regrasped. Additionally, geometric data for objects from a sensor are usually available in the form of point clouds. The problem of computing grasps and regrasps from point-cloud representation of objects for complex manipulation tasks is a key problem in endowing robots with manipulation capabilities beyond pick-and-place. 
In this paper, we formalize the problem of grasping/regrasping for complex manipulation tasks with objects represented by (partial) point clouds and present an algorithm to solve it. We represent a complex manipulation task as a sequence of constant screw motions. Using a manipulation plan skeleton as a sequence of constant screw motions, we use a grasp metric to find graspable regions on the object for every constant screw segment. The overlap of the graspable regions for contiguous screws are then used to determine when and how many times the object needs to be regrasped. We present experimental results on point cloud data collected from RGB-D sensors to illustrate our approach.
\end{abstract}

\section{Introduction}
\label{section:intro}

\begin{figure*}[ht!]
    \centering
    \includegraphics[width=0.8\textwidth]{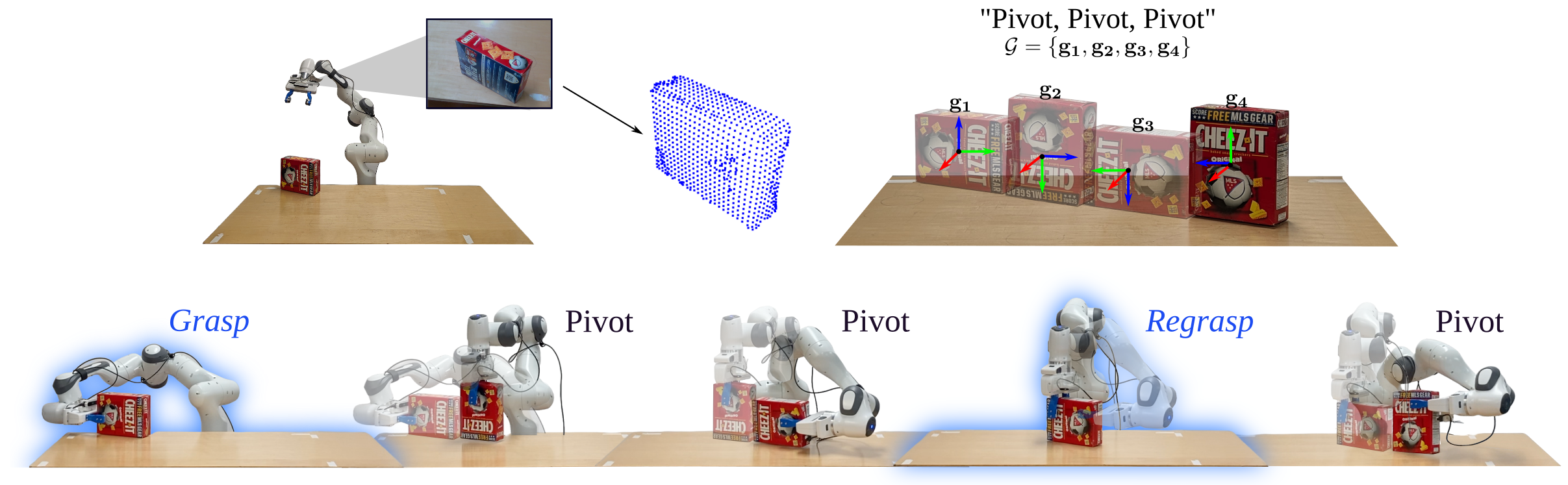}
    \caption{Example task considered in this paper where the robot has to pivot a CheezIt box three times. The partial point cloud of the object is extracted using an eye-in-hand configuration (top left). The sequence of pivoting motions is specified by the motion plan $\mathcal{G}$ (top right). Our approach enables the robot to compute the minimum number of grasps required to execute specified motion indicating that there is a need to \textit{regrasp} the object after the initial two pivoting motions (bottom).}
    \label{fig:overview}
\end{figure*}


Grasping and manipulating an object consists of holding it and moving it through a sequence of configurations. Intuitively, the manner in which an object is held (i.e., the grasp) can affect how easily one can move an object to perform a task. Thus, grasping and motion are intimately related; i.e., how the objects should be moved should determine how they should be grasped. This is especially true for complex manipulation tasks, where there are constraints on the path to be followed by the object (and consequently, the gripper holding the object), and whether the constraints can be satisfied depends on how it is grasped. However, manipulation planning and grasp planning are usually considered separate problems. Grasp planning is usually done without considering the motion that must be performed after grasping. The goal of this paper is to develop a grasp synthesis method that {\em uses information about the motion plan of the object to synthesize grasps. If a single grasp is not suitable for the entire motion plan and the object needs to be regrasped, our algorithm should also explicitly state when an object needs to be regrasped and provide the new grasp}.

We represent the skeleton of the object (or end-effector) manipulation plan in $SE(3)$, the space of all rigid body poses, as a sequence of constant screw segments, i.e., one-parameter subgroups of $SE(3)$~\cite{sarker2020screw}. This is a general way of representing a motion/manipulation plan in the task space, since, as implied by Chasles theorem~\cite{Murray1994}, any path in $SE(3)$ can always be arbitrarily closely approximated by a sequence of constant screw motions~\cite{sarker2020screw,mahalingam2023human}. Additionally, the object geometry is given as a (partial) point cloud, which can be obtained from stereo cameras or RGB-D sensors. Thus, our problem of interest is: {\em Given a (partial) point cloud representation of an object and a manipulation plan skeleton as a sequence of constant screws, compute the minimum number of grasps and the grasping regions needed to manipulate the object according to the plan skeleton.}  

{\bf Illustrative Example}: Consider the task of pivoting a box 
from an initial pose $\bf{g_1}$ to a final pose $\bf{g_4}$ using a sequence of three pivoting motions, each of which is a constant screw motion (see Fig.~\ref{fig:overview}). Intuitively, for constant screw motions ($\bf{g}_1$, $\bf{g}_{2}$) and ($\bf{g}_{2}$, $\bf{g}_{3}$) the same grasp can be used, since the face corresponding to the pivoting edges remains the same. However, the box has to be regrasped for the pivoting motion ($\bf{g}_{3}$, $\bf{g}_{4}$) due to a change in the face corresponding to the pivoting edge. 
This example motivates the need for a grasp synthesis approach that not only considers the motion of the object, but also identifies whether there is a need to \textit{regrasp} the object. Moreover, the approach should be general enough to be extended for other complex manipulation tasks that do not involve contact, such as pouring. 

Pioneered by~\cite{paul1972modelling, tournassoud1987regrasping}, much of the extant literature~\cite{tournassoud1987regrasping, cho2003complete, rohrdanz1997generating, terasaki1998motion, stoeter1999planning, rapela2002planning, lertkultanon2015single, wan2019regrasp} studies regrasping in the context of pick-and-place (or pick-and-insert) operations. The key idea is that regrasping is necessary due to the incompatibility of the grasps for picking and placement (or insertion). We study a complementary aspect of the regrasping problem where the need to regrasp arises in complex manipulation tasks (beyond pick-and-place), due to the motion constraints during manipulation.

{\bf Contributions}: We first formalize the problem of regrasping where object geometry is represented as a (partial) point cloud and the motion constraints are represented using the screw geometry of motion. 
We present an algorithm for computing the grasp regions for contiguous segments of constant screw motions, thereby also computing when to regrasp and how many times regrasping has to be done. Note that in our approach, we do not assume beforehand that a regrasp is necessary. {\em The formalization of the regrasping problem for complex manipulation tasks and the algorithm to solve it using very realistic models of object geometry that are available from modern sensors are the primary contribution of this work}. To validate our approach, we present experimental results on point cloud data obtained from a RGB-D sensor, which shows a success rate of $75\%$. 

\section{Related Work}
\label{section:related_work}

In the context of pick-and-place tasks, regrasping was first studied in~\cite{tournassoud1987regrasping} where the compatibility of a grasp was checked for stable object placement by considering the manipulator kinematic constraints. 
Grasp placement tables were constructed and used to find compatible grasps, and this approach was further extended in~\cite{cho2003complete, rohrdanz1997generating, terasaki1998motion, stoeter1999planning, rapela2002planning, lertkultanon2015single}. Regrasp planning has also been formulated as a graph search problem by constructing a \textit{regrasp} graph. Each regrasp graph has subgraphs corresponding to different stable placements of the objects and each node of the subgraph denotes a grasp~\cite{wan2015improving, wan2017regrasp, wan2019regrasp}. 
Although much work has been done in developing approaches for regrasping to satisfy manipulator kinematics~\cite{tournassoud1987regrasping, cho2003complete, wan2019regrasp} and collision detection~\cite{wan2017regrasp}, it has been mainly applied to object placement tasks~\cite{tournassoud1987regrasping, cho2003complete, wan2019regrasp} and/or object reorienting tasks~\cite{wan2015reorientating, wan2017regrasp, wan2019regrasp} achieved using a sequence of pick-and-place motions. These approaches cannot be applied to complex manipulation tasks like pouring, scooping, or pivoting, where there are constraints on the motion of the object. This is mainly due to the lack of a common mathematical representation for tasks which aids motion planning as well as grasp synthesis. 

In prior work, we have formalized the notion of a task as a constant screw motion about/along a specified unit screw~\cite{fakhari2021computing}, which was then used to formalize the problem of task-oriented grasping as grasping to impart the desired constant screw motion. More recently, we proposed a novel neural network-based algorithm that uses the 3D bounding box of the point cloud along with the task screw as input and computes the 6-DOF end-effector poses to impart the desired motion~\cite{patankar2023task}. Researchers have studied the problem of regrasping, using (partial) point clouds, in the context of pick-and-place tasks~\cite{pavlichenko2019autonomous, cheng2022learning, xu2023predict, xu2023reorient}. However, in this work, we consider complex manipulation tasks that can be represented as a \textit{sequence of constant screw motions}. 


Researchers have studied the problem of \textit{how} the object should be regrasped. Whether the robotic end-effector should maintain continuous contact with the object~\cite{cole1992dynamic, chavan2018regrasping} and reorient/reposition itself to a different grasp pose or if it should completely break contact with the object and regrasp it~\cite{tournassoud1987regrasping, furukawa2006dynamic}. However, the problem we are interested in is \textit{regrasping to satisfy the motion constraints on a specified path}. We assume that the object is placed stably on the surface while regrasping similar to~\cite{tournassoud1987regrasping}. The authors in~\cite{simeonov2021long} study the problem of manipulating an object, using its point cloud, from an initial pose to a final pose using a sequence of primitive skills like \textit{push}, \textit{pull} etc. while maintaining contact with the environment. However, a distinct grasp pose is synthesized for all intermediate object poses. In our work, we represent a task as a sequence of constant screw motions and regrasp the object only if it does not satisfy the motion constraints. Through our results, we show that it may be possible to execute the entire motion using a single grasp.

\section{Preliminaries}
\label{section:prelims}

We now discuss the mathematical preliminaries needed to formalize our problem. The joint space of the robot is the set of all possible joint configurations and is denoted by $\mathcal{J} \subset \mathbb{R}^d$ where $d$ is the number of joints of the robot. The task space of the robot, $\mathcal{T}$, is a subset of $SE(3)$, which is the group of rigid body motions.
Each joint configuration, $\Theta \in \mathcal{J}$, is a $d$-dimensional vector and can be mapped to a pose in the task space using the forward position kinematics. 

\noindent
\textbf{Screw Displacement}: Chasles-Mozzi theorem states that the general Euclidean displacement/motion of a rigid body from the origin $\boldsymbol{I}$ to $\boldsymbol{T} = (\boldsymbol{R},\boldsymbol{p}) \in SE(3)$
can be expressed as a rotation $\theta$ about an axis $\$$, called the \textit{screw axis}, and a translation $d$ along that axis. Plücker coordinates can be used to represent the screw axis by $\boldsymbol{l}$ and $\boldsymbol{m}$, where $\boldsymbol{l} \in \mathbb{R}^3$ is a unit vector that represents the direction of the screw axis $\$$, $\boldsymbol{m} = \boldsymbol{r} \times \boldsymbol{l}$, and $\boldsymbol{r} \in \mathbb{R}^3$ is an arbitrary point on the axis. The pitch of the screw displacement is $h = \theta/d$. Thus, the screw $\mathcal{S}$, corresponding to a screw displacement, can be defined using the parameters $\boldsymbol{l}, \boldsymbol{m}, \theta, h$. Note that a screw can also be represented by using two poses in $SE(3)$. A constant screw motion is a motion where only the magnitude of the screw $\mathcal{S}$ associated with the motion changes, not the pitch nor the axis. Henceforth, we refer to the unit screw corresponding to a task as the \textit{task screw}, $\Tilde{\mathcal{S}} =(\boldsymbol{l}, \boldsymbol{m}, h)$.




\section{Problem Formulation }
\label{section:prob_formulation}


Let $\mathcal{G} = \{{\bf g}_1, {\bf g}_2, {\bf g}_3, \cdots, {\bf g}_k \mid  {\bf g}_i \in \hspace{1mm} $\textit{SE}(3)$,  i=1, \dots, k  \}$ be a motion plan of an object, grasped by the robot, such that the pair (${\bf g}_i$, ${\bf g}_{i+1}$), consisting of two consecutive object poses, represents a constant screw motion. Thus, $\mathcal{G}$ will have $k-1$ constant screw segments. Let $\Tilde{\mathcal{S}_i} = (\bm{l}_i, \bm{m}_i, h_i)$ be the task screw corresponding to the constant screw motion (${\bf g}_i$, ${\bf g}_{i+1}$). Note that this is a general way to represent motion plans in the task space $\mathcal{T}$ not only for complex manipulation tasks like pouring~\cite{mahalingam2023human} but also for tasks that exploit contact with the environment like pivoting~\cite{fakhari2021motion}.
Let $\mathcal{O}_c$ be the point set representation of the object being manipulated. 
Let $\mathcal{O}_{{\bf g}_1} \subset \mathcal{O}_c$ be the point cloud representation of the object available to us in the initial pose ${\bf g}_1 \in \mathcal{G}$. \textit{
Given the point cloud representation of the object $\mathcal{O}_{{\bf g}_1}$ and a sequence of constant screw segments, $\mathcal{G}$, compute the minimum number of grasps required for successfully executing the motion plan specified by $\mathcal{G}$.} In our work, we define \textit{regrasping} as the need to change the grasp pose between any two consecutive constant screw motions, (${\bf g}_i$, ${\bf g}_{i+1}$) and (${\bf g}_{i+1}$, ${\bf g}_{i+2}$), for the successful execution of the motion plan $\mathcal{G}$.

\begin{figure*}[ht!]
    \centering    \includegraphics[width=0.75\textwidth]{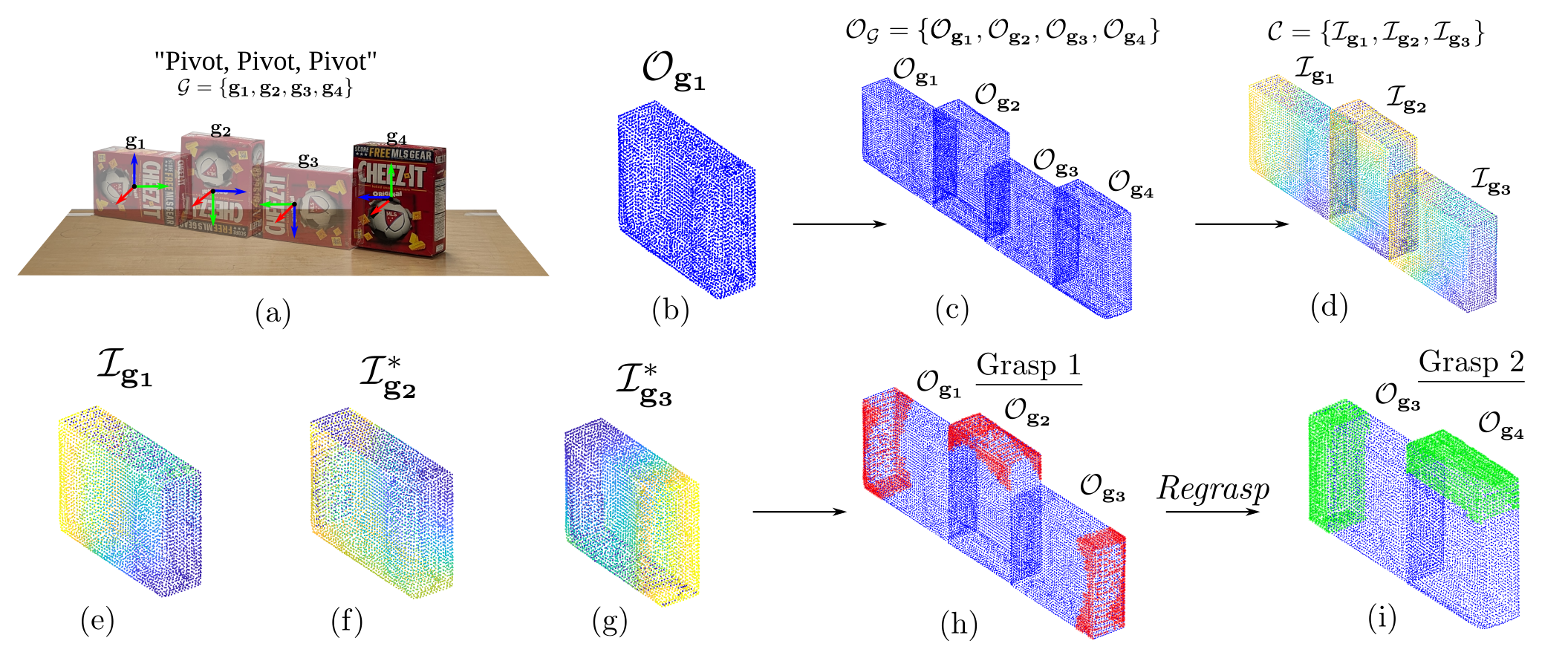}
    \caption{Solution overview with pivoting a CheezIt box three times. Using the \textit{complete} point cloud description of the CheezIt box available at the initial pose $\mathcal{O}_{{\bf g}_1}$ and the motion plan $\mathcal{G}$ we compute the set of point cloud representations $\mathcal{O}_{\mathcal{G}}$ and the set of ideal grasping regions $\mathcal{C}$, visualized in yellow (a-d). By sequentially comparing the computed grasping regions, in the appropriate reference frame, we see that we need a minimum of $2$ grasps (e-g); Grasp 1 (red) for the first two pivots and Grasp 2 (green) for the last pivot (h-i). }
    \label{fig:solution_approach}
    \vskip10pt
\end{figure*}


\noindent
{\bf Solution Approach Overview: }
For the given object point cloud, $\mathcal{O}_{{\bf g}_1}$, and the motion plan $\mathcal{G}$, we first compute the point clouds at each of the poses ${\bf g}_i \in \mathcal{G}$ by transforming the point cloud $\mathcal{O}_{{\bf g}_1}$. We denote the set of all object point clouds corresponding to each object pose ${\bf g}_i \in \mathcal{G}$ as $\mathcal{O}_{\mathcal{G}} = \{ \mathcal{O}_{{\bf g}_1}, \mathcal{O}_{{\bf g}_2}, \mathcal{O}_{{\bf g}_3},  \cdots, \mathcal{O}_{{\bf g}_{k}} \mid \mathcal{O}_{{\bf g}_i} \subseteq \mathcal{O}_c, i=1, \dots, k   \}$.
We then compute the grasping region $\mathcal{I}_{{\bf g}_i} \subseteq \mathcal{O}_{{\bf g}_i}$ corresponding to each constant screw motion (${\bf g}_i$, ${\bf g}_{i+1}$) in $\mathcal{G}$ (using~\cite{patankar2023task}). By sequentially comparing each grasping region $\mathcal{I}_{{\bf g}_i}$ with all its subsequent grasping regions using a proposed score measure $\gamma$, we compute whether there is a need to regrasp the object between any two consecutive constant screw motions (${\bf g}_i$, ${\bf g}_{i+1}$) and (${\bf g}_{i+1}$, ${\bf g}_{i+2}$) in $\mathcal{G}$ or whether the same grasp can be used to execute both.

\section{Synthesizing Task-Dependent (Re)Grasps}
\label{section:solution_approach}




We will now discuss our solution approach for grasp synthesis with re-grasping for a given manipulation task. 
Algorithm~\ref{regrasping-algorithm} summarizes our approach. We will use the running example of manipulating a box by pivoting (see Fig~\ref{fig:solution_approach}) for illustrating key steps in Algorithm~\ref{regrasping-algorithm}. For ease of exposition, we use the \textit{complete} point cloud of the CheezIt box available as part of the YCB dataset~\cite{calli2015ycb}. 

For completeness, we briefly review our task-dependent grasp metric $\eta$ used for computing the grasping regions on object point clouds. The grasp metric $\eta$ is the maximum magnitude of wrench (force-moment pair) that can be exerted about the screw axis $\$$, corresponding to a task screw $\Tilde{S}$, while satisying the friction cone constraints at the object-robot and object-environment contacts~\cite{fakhari2021computing}. 
We also consider the effect of external wrenches acting on the object during the motion. Thus, $\eta$ allows us to evaluate the feasibility of object-robot contact locations to impart the desired constant screw motion, while considering the physics and geometric constraints of the motion. 


\subsection{Computing Grasping Region from Point Cloud Data }

In Algorithm~\ref{regrasping-algorithm}, the input is the motion plan $\mathcal{G}$,
a point cloud of the object $\mathcal{O}_{{\bf g}_1}$ at the initial pose ${\bf g}_1$ and a threshold values for score metric $\gamma_{\text{th}}$ and grasp metric $\eta_{\text{th}}$. Let ${\bf p}_{ji} \in \mathcal{O}_{{\bf g}_i}$ be the homogeneous coordinates of the position of the point $j$ on the object at pose ${\bf g}_i$ expressed in the frame ${\bf g}_i$. At pose ${\bf g}_i$, the coordinates of point $j$ in the body frame of the object at the initial pose ${\bf g}_1$ is ${\bf p}_{ji} = {\bf g}_1^{-1}{\bf g}_i{\bf p}_{j1}$. Using this to transform each point, for each pose ${\bf g}_i \in \mathcal{G} $, we can compute the corresponding point cloud $\mathcal{O}_{{\bf g}_i}$. This is implemented using the function \texttt{transformPointCloud}() in Algorithm~\ref{regrasping-algorithm}.
%
The function \texttt{computeMetric}() is used to compute the set of all grasping regions denoted by $\mathcal{C} = \{ \mathcal{I}_{{\bf g}_1}, \mathcal{I}_{{\bf g}_2}, \mathcal{I}_{{\bf g}_3},  \cdots, \mathcal{I}_{{\bf g}_{k-1}} \mid \mathcal{I}_{{\bf g}_i} \subseteq \mathcal{O}_{{\bf g}_i}, i=1, \dots, k-1   \}$. This function implements our task-oriented grasp synthesis approach in~\cite{patankar2023task}.
Note that each $\mathcal{I}_{{\bf g}_i} \in \mathcal{C}$ is computed with respect to the object's local reference frame ${\bf g}_i \in \mathcal{G}$ and is based on a threshold value $\eta_{\text{th}}$ of our metric~\cite{fakhari2021computing}. Since the task of pivoting consists of three constant screw motions, 
the corresponding set $\mathcal{C}$ consists three elements $\mathcal{I}_{{\bf g}_1}, \mathcal{I}_{{\bf g}_3}$ and $\mathcal{I}_{{\bf g}_3}$ (depicted in yellow) as shown in Fig~\ref{fig:solution_approach}-d. 

\subsection{Formalizing Regrasping using Grasping Regions} 


We now formalize the regrasping problem using the notion of grasping regions.
Let the index set of the elements of $\mathcal{C}$, be denoted by $\mathcal{J} =  \{ 1, 2, \cdots, k-1\}$. A sequential partition of the index set $\mathcal{J}$ into $\alpha$ components is: $\mathcal{J} = \mathcal{J}_1 \cup \mathcal{J}_2 \cup \cdots \cup \mathcal{J}_{\alpha}$ such that $\mathcal{J}_u \cap \mathcal{J}_v = \varnothing$ where $u, v \in \{1, 2, \cdots, \alpha\}, u \neq v, u < v$. The index set partition induces a partition on $\mathcal{C}$ into subsets $\mathcal{C}_u$ such that $\mathcal{C}_u = \{ \mathcal{I}_{{\bf g}_j} \vert j \in \mathcal{J}_u, \mathcal{I}_{{\bf g}_j} \in \mathcal{C}\}$. Thus, our goal of \textit{computing the minimum number of grasps required for the task specified by $\mathcal{G}$} 
can be expressed as
\vskip-8pt
\begin{equation}
\begin{aligned}
& {\text{minimize} } & & \alpha \\
& \text{such that}  & & \mathcal{C} = \mathcal{C}_1 \cup \mathcal{C}_2 \cup \cdots \cup \mathcal{C}_{\alpha}, \\
&&& \bigcap\limits_{j \in \mathcal{J}_u} \mathcal{I}_{{\bf g}_{j}} \neq \varnothing , \forall u = 1, 2, \cdots \alpha, 
\end{aligned}
\label{equation:updated_optimization}
\end{equation}

The above formulation can be used to partition the set $\mathcal{C}$ into smaller subsets $\mathcal{C}_u$ whose elements are the grasping regions $\mathcal{I}_{{\bf g}_i} \in \mathcal{C}$ with non-zero intersection. The total number of such subsets, i.e., $\alpha$ indicates the number of \textit{different} grasps required to perform the task given by $\mathcal{G}$. The number of regrasping operations is thus $\alpha - 1$. 
In Algorithm~\ref{regrasping-algorithm} the set of all such subsets $\mathcal{C}_u$ is denoted by $\mathcal{Z}$ and thus $|\mathcal{Z}| = \alpha$. 




\SetCommentSty{CommentFont}
\SetKwInput{KwInput}{Input}
\SetKwInput{KwOutput}{Output}
\begin{algorithm}[t!] 
\DontPrintSemicolon
  \KwInput{Motion plan skeleton $\mathcal{G}$; Point cloud of the object at initial pose $\mathcal{O}_{{\bf g}_1}$; Thresholds, $\gamma_{\text{th}}$, $\eta_{\text{th}}$ value for the score and grasp metric.}
  
  \KwOutput{Number of distinct grasps $\alpha$}
  
  
  $\mathcal{O}_{\mathcal{G}} \gets$ \SetKwFunction{FMain}{transformPointCloud}\SetKwProg{Fn}{:}{}
  \Fn{\FMain{$\mathcal{O}_{{\bf g}_1}, \mathcal{G}$}}
    \SetKwFunction{FMain}

  $\mathcal{C} \gets$ \SetKwFunction{FMain}
  {computeMetric}\SetKwProg{Fn}{:}{}
  \Fn{\FMain{$\mathcal{O}_{\mathcal{G}}, \mathcal{G}$, $\eta_{\text{th}}$}}
    \SetKwFunction{FMain}
  
    
    Initialize the set $\mathcal{Z} = \{ \}$ and $i = 1$
    
    \While {$i < k-1$}
    {
        
        Initialize $\mathcal{C}_u = \{ \mathcal{I}_{{\bf g}_i}\}$

        \For{$j = 2, \cdots k-1$}
        {

        $\mathcal{I}^*_{{\bf g}_{j}} \gets$ \SetKwFunction{FMain}{transformRegion}
    \SetKwProg{Fn}{:}{}
    \Fn{\FMain{${\bf g}_{j},{\bf g}_{i},\mathcal{I}_{{\bf g}_{j}}$}}

       $\mathcal{C}_u \gets$  \SetKwFunction{FMain}{append}
    \SetKwProg{Fn}{:}{}
    \Fn{\FMain{$\mathcal{I}^{*}_{{\bf g}_{j}}$}}

        $\gamma \gets$ \SetKwFunction{FMain}{computeScore}
    \SetKwProg{Fn}{:}{}
    \Fn{\FMain{$\mathcal{C}_u$}}


        \If{$\gamma < \gamma_{\text{th}}$}
        {

            $\quad$ $\mathcal{Z} \gets$ \SetKwFunction{FMain}{append}
    \SetKwProg{Fn}{:}{}
    \Fn{\FMain{$\mathcal{C}_u$}}
    
            
            $\quad$ Set $i = j$
            $\quad$ \textbf{break}
        }
        
        \ElseIf{j = k-1}
        {

            $\quad$ $\mathcal{Z} \gets$\SetKwFunction{FMain}{append}
    \SetKwProg{Fn}{:}{}
    \Fn{\FMain{$\mathcal{C}_u$}}
            
            $\quad$ \textbf{break}
        }

        \Else{
            $\quad$ \textbf{continue}
        }
        
        }        
    }
    $\alpha = \lvert \mathcal{Z} \rvert$
  
\caption{Compute $\alpha$ and $\mathcal{Z}$}
\label{regrasping-algorithm}
\end{algorithm}
\setlength{\textfloatsep}{0pt}

To compute the subsets $\mathcal{C}_u$, we sequentially compare the extent of overlap between a particular ideal grasping region $\mathcal{I}_{{\bf g}_i}  \in \mathcal{C}$ with all subsequent ideal grasping regions $\mathcal{I}_{{\bf g}_{j}} \in \mathcal{C}$, $\forall j = i+1, i+2, \cdots k-1$ using the score measure $\gamma$ (described in the next section). Recall that $\mathcal{I}_{{\bf g}_i}$ is the grasping region associated with the constant screw motion (${\bf g}_i$, ${\bf g}_{i+1}$), expressed with respect to the frame ${\bf g}_i$. Thus, to compare two ideal grasping regions, they need to be expressed in the same reference frame. The transformed grasping region is denoted by $\mathcal{I}^{*}_{{\bf g}_{j}}$ as shown in Algorithm~\ref{regrasping-algorithm} line 7. For the task of pivoting a CheezIt box three times with $\mathcal{C} = \{\mathcal{I}_{{\bf g}_1}, \mathcal{I}_{{\bf g}_2},\mathcal{I}_{{\bf g}_3}\}$, we initialize the set $\mathcal{C}_u$ with $\mathcal{I}_{{\bf g}_1}$ as per line 5 in Algorithm~\ref{regrasping-algorithm}. Next, we transform the grasping regions $\mathcal{I}_{{\bf g}_2}$ and $\mathcal{I}_{{\bf g}_3}$ with respect to ${\bf g}_1$. The transformed grasping regions, $\mathcal{I}^*_{{\bf g}_2}$ and $\mathcal{I}^*_{{\bf g}_3}$, are shown in Fig~\ref{fig:solution_approach}-(f) and Fig~\ref{fig:solution_approach}-(g) respectively.  


\SetCommentSty{CommentFont}
\SetKwInput{KwInput}{Input}
\SetKwInput{KwOutput}{Output}
\begin{algorithm}[t!] 
\DontPrintSemicolon
  \KwInput{The set $\mathcal{C}_u$} 
  \KwOutput{The score value $\gamma$}
  
    \SetKwFunction{FMain}{computeScore}
    \SetKwProg{Fn}{def}{:}{}
    \Fn{\FMain{$\mathcal{C}_u$}}{
        
        Initialize the list $\Gamma = [\hspace{1mm}]$ of computed scores.
        
        $\mathcal{C}_u = \{ \mathcal{I}_{g_1}, \mathcal{I}_{g_2}, \cdots, \mathcal{I}_{g_n}\}$

        $\mathrm{I} = \bigcap\limits_{ i = 1}^{n} \mathcal{C}_u$

    \For{$i = 1, 2 , \cdots, n$}{

    $\gamma_i = \frac{\lvert \mathrm{I} \rvert}{\lvert \mathcal{I}_{g_i} \rvert}$

    $\Gamma \gets$\SetKwFunction{FMain}{append}
    \SetKwProg{Fn}{:}{}
    \Fn{\FMain{$\gamma_i$}}

  }
  $\gamma \gets $\SetKwFunction{FMain}{min}
    \SetKwProg{Fn}{:}{}
    \Fn{\FMain{$\Gamma$}}

  \textbf{return} $ \gamma; $
}
  
\caption{Compute the score $\gamma$} 
\label{compute-gamma-algorithm}
\end{algorithm}
\setlength{\textfloatsep}{0pt}

\begin{figure*}[ht!]
    \centering
    \includegraphics[width=0.85\textwidth]{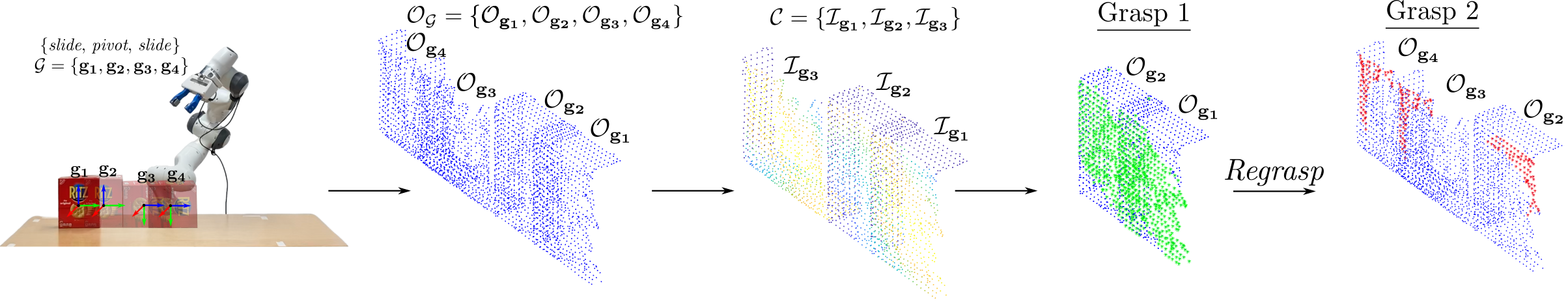}
    \caption{Result where two grasps, i.e., one regrasping suffice to perform the task given by the plan skeleton $\{slide, pivot, slide\}$ using the partial point cloud of a Ritz cracker box. One grasp (green) for the first sliding motion and one (red) for subsequent pivoting and sliding.}
    \label{fig:results_1}
\end{figure*}

\subsection{Comparing Grasping Regions for Regrasping}
We now describe our approach for computing the score $\gamma$ to compare the extent of overlap between a grasping region $\mathcal{I}_{{\bf g}_i}$ and all its subsequent grasping regions $\mathcal{I}_{{\bf g}_{j}} \in \mathcal{C}$, $\forall j = i+1, i+2, \cdots k-1$. We also discuss how the score $\gamma$ is used to compute the minimum number of grasps $\alpha$. To compute $\gamma $ for a set $\mathcal{C}_u \subset \mathcal{C}$ with $\mathcal{C}_u = \{ \mathcal{I}_{{\bf g}_1},  \cdots, \mathcal{I}_{{\bf g}_{n}}\}$, we define $\gamma_i$ for each $\mathcal{I}_{{\bf g}_i} \in \mathcal{C}_u$ as $\gamma_i = \lvert \mathrm{I} \rvert / \lvert \mathcal{I}_{{\bf g}_i} \rvert$, where $\lvert \cdot \rvert$ denotes the cardinality of a set and $\mathrm{I} = \bigcap\limits_{ i = 1}^{n} \mathcal{C}_u$ is the common region of all elements of the set $\mathcal{C}_u$. Thus, {\em $\gamma_i$ is an estimate for the fraction of the area of $\mathcal{I}_{{\bf g}_i}$ that is good for grasping for the sequence of $n$ constant screw motion segments starting at ${\bf g}_i$}.
The score $\gamma$ is the minimum of all the $\gamma_i$'s. If $\gamma \geq \gamma_{\text{th}}$, then a single grasp can be used across all the $n$ motion segments corresponding to the grasping regions in $\mathcal{C}_u$ and the grasp contacts will lie in $\mathrm{I}$.
Although a grasp exists for all the $n$ motion segments if $\gamma > 0$, the thresholding ensures robustness of the grasp by requiring that a significant portion of the graspable region is common across all the motion segments. Algorithm~\ref{compute-gamma-algorithm} summarizes the computation of $\gamma$.

Revisiting our example in Fig.~\ref{fig:solution_approach}, 
for the first iteration, the set $\mathcal{C}_u = \{ \mathcal{I}_{{\bf g}_1}, \mathcal{I}^*_{{\bf g}_2}\}$.  We set the threshold value as $\gamma_{\text{th}} = 0.25$ and for the set $\mathcal{C}_u = \{ \mathcal{I}_{{\bf g}_1}, \mathcal{I}^*_{{\bf g}_2}\}$ we get $\Gamma = [0.583, 0.926]$ and the score $\gamma = 0.583$ as per lines $7$ and $8$ in Algorithm~\ref{compute-gamma-algorithm}. Note that the actual values of $\gamma_i$ depend on the threshold $\eta_{\text{th}}$, for our task-dependent grasp metric $\eta$~\cite{fakhari2021computing},  used while computing the ideal grasping regions in $\mathcal{C}$~\cite{patankar2023task}. In this example, we set $\eta_{\text{th}} = 0.75$ and selecting a different value for $\eta_{\text{th}}$ only scales $\gamma_i$. For the set $\mathcal{C}_u = \{ \mathcal{I}_{{\bf g}_1}, \mathcal{I}^*_{{\bf g}_2}\}$ since $\gamma \geq \gamma_{\text{th}}$ the same grasp can be used to execute the constant screw motions (${\bf g}_1$, ${\bf g}_{2}$) and (${\bf g}_{2}$, ${\bf g}_{3}$) as shown in Fig.~\ref{fig:solution_approach}-(h). 

\vspace{-0.5mm}

If $\gamma < \gamma_{\text{th}}$ after including a particular grasping region $\mathcal{I}^{*}_{{\bf g}_{j}}$ in $\mathcal{C}_u$ then we exit the loop as depicted in line 10 of Algorithm~\ref{regrasping-algorithm}. For our pivoting example, this is observed after we include the third transformed grasping region $\mathcal{I}^{*}_{{\bf g}_{j}} = \mathcal{I}^{*}_{{\bf g}_{3}}$ in $\mathcal{C}_u = \{ \mathcal{I}_{{\bf g}_1}, \mathcal{I}^*_{{\bf g}_2}\}$. After including $\mathcal{I}^{*}_{{\bf g}_{3}}$ $,\Gamma = [0, 0, 0]$ which indicates that no intersection exists between $\mathcal{I}_{{\bf g}_{1}}$, $\mathcal{I}^{*}_{{\bf g}_2}$ and $\mathcal{I}^*_{{\bf g}_3}$. In such situations, the set $\mathcal{C}_u$ contains all the transformed grasping regions till $\mathcal{I}^{*}_{{\bf g}_{j-1}}$ along with $\mathcal{I}_{{\bf g}_{i}}$ and is essentially one of the subsets of $\mathcal{C}$ which we are interested in computing.  The set $\mathcal{C}_u$ is then stored in $\mathcal{Z}$ this process is repeated and the steps 9-16 in Algorithm~\ref{regrasping-algorithm} indicate the same. After iterating over all the grasping regions $\mathcal{I}_{{\bf g}_i} \in \mathcal{C}$ and computing the subsets $\mathcal{C}_u$ as described above the cardinality of the resultant set $\mathcal{Z}$ indicates the \textit{minimum} number of grasps $\alpha$ required for executing the sequence of constant screw motions $\mathcal{G}$. The number of regrasping operations required is then $\alpha - 1$. Therefore for our example $\mathcal{C} = \{\mathcal{I}_{{\bf g}_1}, \mathcal{I}_{{\bf g}_2},\mathcal{I}_{{\bf g}_3}\}$ we get $\mathcal{Z} = \{\{ \mathcal{I}_{{\bf g}_1}, \mathcal{I}^*_{{\bf g}_2}\}\{\mathcal{I}^*_{{\bf g}_3}\}\}$ and $\alpha = 2$ indicating that two grasps are required as shown in Fig~\ref{fig:solution_approach}-(h) and Fig~\ref{fig:solution_approach}-(i).

\section{Experimental Evaluation}
\label{section:eval}

\begin{figure*}[ht!]
    \centering    \includegraphics[width=0.8\textwidth]{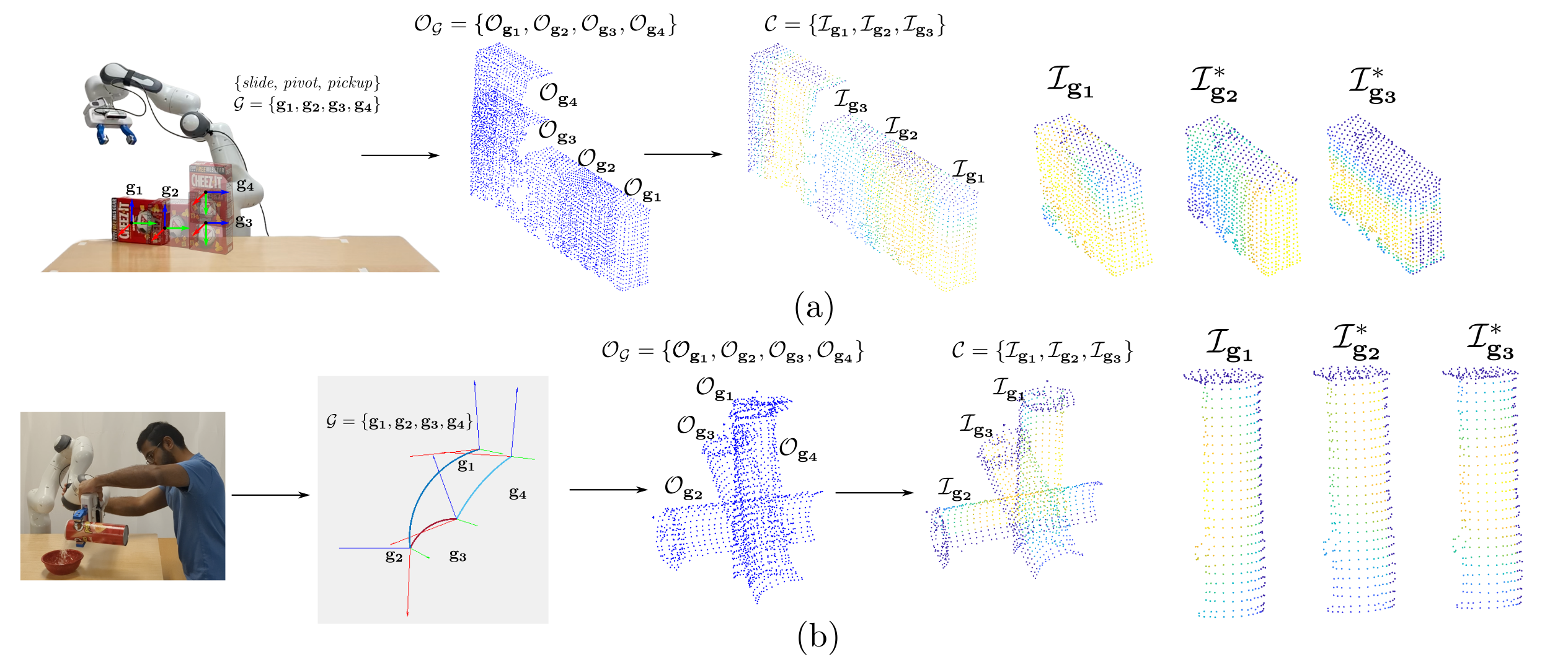}
    \caption{Results where a single grasp is sufficient to perform the task (a) Plan skeleton $\{slide, pivot, pickup\}$ using the partial point cloud of a CheezIt box. (b) Pouring using the partial point cloud of the Pringles container. The first two images show the acquisition of a kinesthetic demonstration and extracted task constraints as a sequence of constant screw motions.} 
    \label{fig:results_2}
\end{figure*}

We evaluate the performance of our approach by
(a) computing the minimum number of grasps for a given manipulation plan, using object data obtained from an RGBD camera, and (b) performing real world experiments with a robot where the objective is to impart the desired sequence of constant screw motions while regrasping the object if necessary. Please note that there are no existing algorithm that is capable of synthesizing grasps for the type of tasks that we consider using point cloud representation of objects. So there are no baselines to compare against and we hope that our approach forms a baseline for future work.
We consider tasks where the motion constraints on the object being manipulated come from contact with the environment, as well as tasks like pouring where there are motion constraints come from the nature of the task. We show that for a given sequence of constant screw motions, our approach can compute the minimum number of grasps required to execute the motion while satisfying the related constraints. 


\noindent
{\bf Experimental Setup}:
We used an Intel Realsense D415 camera mounted on a Franka Emika Panda using the same perception pipeline, to extract object point clouds, as described in~\cite{patankar2023task}. Point cloud processing and 3D bounding box computation are performed using off-the-shelf algorithms. For any object, the vertices of its 3D bounding box $\mathcal{B}$ are used to compute the object pose ${\bf g}_1 \in SE(3)$, with respect to the robot's base frame. Thus, the point cloud $\mathcal{O}$ is available to us in pose ${\bf g}_1$ and we denote it by $\mathcal{O}_{{\bf g}_1}$. Additionally, 
the task screw $\Tilde{S}$ is also known to us in the frame ${\bf g}_1$. 

\subsection{Evaluation using Real Point Clouds in Simulation}
We use three plan skeletons: $\{pivot, pivot, pivot\}$, $\{slide, pivot, slide\}$ and $\{slide, pivot, pickup\}$ as a high-level representation of the motion plan $\mathcal{G}$. Each element of a skeleton represents a \textit{single} constant screw motion about/along the associated screw, $\mathcal{S}_i = (\bm{l}_i, \bm{m}_i, \theta_i, h_i)$.
Note that $pivoting$ represents pure rotational motion ($h_i = 0$) about the screw axis $\$$ whereas both $sliding$ and $pickup$ represent pure translation ($h_i = \infty$) along their respective screw axis $\$$. Please note that although our plan skeleton consists of 3 constant screw motions, our proposed approach for regrasping can be used more generally for motion plans consisting of $k$ constant screw motions. 

We collect point cloud data for $4$ objects from $4$ different poses each. For each object-pose pair we compute the minimum number of distinct grasps $\alpha$ and the corresponding subsets $\mathcal{Z}$ required to execute the motion plans associated with the 3 plan skeletons. This gives us a total of 48 trials. 
For {\em pivoting}, we consider two object-environment contacts one at each vertex of the pivoting edge, and for sliding, we consider a single object-environment contact located at the center of the bottom face of the bounding box. The task-dependent grasp metric $\eta$ for pivoting is the maximum magnitude of moment that can be generated about the screw axis $\$$. For the $sliding$ and $pickup$ motions, $\eta$ is the maximum magnitude of force applied along the screw axis $\$$.   Our optimization formulation is a second order cone program and has been implemented using CVXPY~\cite{diamond2016cvxpy} in Python. 




\noindent
\textbf{Results and Discussion: } 
Sample results 
are shown in Fig~\ref{fig:results_1}-(a), Fig~\ref{fig:results_1}-(b) and Fig~\ref{fig:results_2}-(a) 
using the partial point cloud of a CheezIt box.
For all $4$ objects, using the plan skeletons $\{pivot, pivot, pivot\}$ and $\{slide, pivot, slide\}$ gives $\alpha = 2$, i.e., at least $2$ grasps are required to execute the corresponding motion. However, the corresponding subsets $\mathcal{C}_u$ may vary. For example, depending on the initial pose of the CheezIt box with the skeleton $\{pivot, pivot, pivot\}$, the set $\mathcal{Z}$ may be either $\{\{\mathcal{I}_{{\bf g}_1}, \mathcal{I}_{{\bf g}_2}\},\{\mathcal{I}_{{\bf g}_3}\} \}$ or $\{\{ \mathcal{I}_{{\bf g}_1}\},\{\mathcal{I}_{{\bf g}_2}, \mathcal{I}_{{\bf g}_3}\} \}$ as seen in Fig~\ref{fig:solution_approach} and Fig~\ref{fig:results_1} respectively. For the task of $\{slide, pivot, pickup\}$ using a CheezIt box or a Domino sugar box, depending on the initial pose, the minimum number of distinct grasps $\alpha $ is either $1$ or $2$. Whereas  
for the Ritz cracker and Pringles container at least $2$ grasps are required. The Fig.~\ref{fig:results_2}-(a) shows the results for $\{slide, pivot, pickup\}$ using a partial point cloud of the CheezIt box with $\alpha = 1$, $\mathcal{Z} = \{ \{\ \mathcal{I}_{{\bf g}_1}, \mathcal{I}_{{\bf g}_2},\mathcal{I}_{{\bf g}_3}\}\}$ and $\Gamma = [0.543, 0.756, 0.6238]$ indicating that a single grasp can be used to execute the motion. It takes approximately 30 seconds to compute $\alpha$ for a single plan on an Intel $i7$ processor with $16$ GB RAM.

\subsection{Grasp Synthesis with Motion Planning using a Robot}

We also evaluate our approach for task-oriented grasp synthesis with regrasping by executing the motion associated with the three plan skeletons described above as well as for the task of pouring out the contents from a container into a bowl using a robot (please see accompanying video). We collect point cloud data corresponding to the CheezIt and Ritz cracker boxes from two different poses each. Next for each point cloud $\mathcal{O}_{{\bf g}_1}$, using the three plan skeletons described previously, the minimum number of grasps $\alpha$ and the set $\mathcal{Z}$ containing the different subsets of $\mathcal{C}$ are computed as described in Section~\ref{section:solution_approach}. Similarly for the task of pouring we collect point cloud data from four different locations for two objects namely the Spam and the Pringles container. The task-related constraints are extracted as a sequence of constant screw motions from two kinesthetic demonstrations of pouring (one for each object) as described in~\cite{mahalingam2023human}. This extracted sequence of constant screw motions is the motion plan $\mathcal{G}$ and is used to compute $\alpha$ and $\mathcal{Z}$ as shown in Fig~\ref{fig:results_2}-(b). Note that for the task of pouring, we use a pretrained neural network~\cite{patankar2023task} for computing the grasping region corresponding to an object point cloud.


For all trials, the intersection $\mathrm{I}$ of each subset $\mathcal{C}_u$ in  $\mathcal{Z}$ is used to compute a set of end-effector poses 
using the bounding-box based algorithm described in~\cite{patankar2023task}. Given the sequence of computed end-effector poses $\mathcal{E}$ a ScLERP-based motion planner with Jacobian pseudoinverse~\cite{sarker2020screw} is used to compute the motion plan in the joint space. 




\captionsetup{belowskip=-12pt}
\begin{table}[ht!]
\centering
\begin{tabular}{|c|c|c|c|c|}
\cline{1-5}
\textbf{Object} & \textbf{Trials}  & \textbf{Successful} &  max $\alpha$ &  min $\alpha$ \\ \cline{1-5}
CheezIt Box & 6 & 4 & 2 & 1 \\ \cline{1-5}
Ritz Cracker Box & 6 & 4 & 2 & 2 \\ \cline{1-5}
Spam Container (Pouring) & 4 & 4 & 1 & 1 \\ \cline{1-5}
Pringles (Pouring) & 4 & 3 & 1 & 1 \\ \cline{1-5}
\end{tabular}
\caption{Experimental Results for Grasping and Motion Planning using Franka Emika Panda}
\label{Grasping_with_Motion_Planning_Robot}
\end{table}

\noindent
\textbf{Results and Discussion: } The results of our experiments with the robot are shown in Table~\ref{Grasping_with_Motion_Planning_Robot}. The first column shows the objects used. The second column shows the total number of trials and the third column shows the number of successful trials. We consider a trial to be unsuccessful if the motion plan generation fails due to hitting the joint limits. The last two columns show the maximum and minimum values of $\alpha$. The first two rows show the results for executing all three plan skeletons using partial point clouds of the CheezIt and Ritz cracker box. The last two rows show the results for executing the pouring task using the Spam and Pringles container. Note that a single grasp ($\alpha = 1$) was required for the plan $\{slide, pivot, pickup\}$ using the CheezIt box, as well as for pouring from both the Spam and Pringles container. A sample result for the Pringles container is shown in Fig~\ref{fig:results_2}-(b), where $\alpha = 1$ and $\gamma = 0.890$.

Although we were able to compute the appropriate grasping regions to satisfy task constraints, motion plan generation failed for the CheezIt box while generating the plan for the skeletons $\{slide, pivot, slide\}$ and $\{slide, pivot, pickup\}$. For the Ritz cracker box, failures were observed for $\{pivot, pivot, pivot\}$ and $\{slide, pivot, pickup\}$. Failures were also observed during pouring. As stated above, these failures were due to the manipulator hitting the joint limits. This is to be expected since we have not considered the joint space constraints of the manipulator in our approach, which we plan to incorporate in future work.


\section{Conclusion and Future Work}
\label{section:conclusion}
In this paper, we presented a novel approach to grasp (and regrasp) synthesis for complex manipulation tasks using a point cloud model for the geometry of objects. The key aspect of our paper is to consider the path constraints on the end-effector (gripper/object) during the motion. Using a formal representation of the path constraints as a sequence of constant screw motions and a grasp metric to compute the grasping region to move an object along a constant screw, we present an algorithm to compute the minimum number of grasps required to perform a manipulation task instance and the corresponding grasps. We evaluated our algorithm on real point-cloud data (with self-occlusion). We also presented experimental results on the execution of the manipulation plans using the grasps that we obtained, showing a success rate of about $75\%$. 
The (re)grasp synthesis problem that we considered took into consideration the constraints on the motion of the object, but did not consider the joint limits of the manipulator. In future work, we plan to incorporate the joint limit constraints of the manipulator. In addition, collision constraints of the gripper/manipulator with the objects will also be incorporated.

\bibliographystyle{IEEEtran}
\bibliography{references}

\end{document}